\pdfoutput=1
\documentclass[11pt,a4paper,twocolumn]{article}

\usepackage[utf8]{inputenc}
\usepackage[english]{babel}
\usepackage{csquotes}
\usepackage{amsmath}
\usepackage{textcomp}
\usepackage{preprint}

\usepackage[style=apa, backend=biber, uniquename=false, sortcites=true, uniquelist=false]{biblatex}
\hypersetup{
  pdftitle  = {Concept Representations in Vector Space: Toward an Interdisciplinary Study of Meaning Variation},
  pdfauthor = {Simon Josef Durstewitz}
}

\title{Concept Representations in Vector Space:\\
Toward an Interdisciplinary Study of Meaning Variation}

\authors{%
  {\large\textbf{Simon Josef Durstewitz}}\\[0.6em]
  {\small Copenhagen Center for Social Data Science (SODAS), University of Copenhagen, Denmark}\\[0.3em]
  {\small\href{mailto:cjm902@sodas.ku.dk}{\texttt{cjm902@sodas.ku.dk}}}%
}

\newcommand{\coderepo}{\href{https://github.com/simon-josef/conrep}{github.com/simon-josef/conrep}}

\preprintnote{Preprint. Correspondence: Simon Josef Durstewitz (\href{mailto:simon.durstewitz@gmail.com}{simon.durstewitz@gmail.com}). Code: \coderepo.}

\begin{document}

\maketitle

\begin{abstract}
An increasing number of disciplines recognize the explanatory potential of meaning-bearing mental structures through which individuals interpret the world and guide their behavior. Concept representations constitute a core class of these structures, emerging from individuals' experiences with concept referents and encoding and organizing their understanding of them. Recognizing a private level of meaning, where concepts are not objectively defined by rules in the world but constitute subjective structures in individuals' minds, raises a fundamental question: do different individuals actually mean the same thing when they invoke the same concept? As latent structures, concept representations are difficult to access and model empirically. The present study introduces a measurement framework that makes such representations empirically accessible through textual traces elicited under standardized conditions. Using transformer-based sentence embeddings, the framework maps standardized responses elicited for concepts into a high-dimensional shared vector space, opening concept representations to the systematic analysis of interindividual variation. It develops three analytical avenues: (1) comparing representational sharedness across concepts, (2) identifying variables that correlate with representational variation, and (3) examining how focal actors or subpopulations deviate from a reference population. Each avenue is illustrated through an empirical proof-of-concept drawing on a large-scale free-association dataset. Together, the demonstrations show how researchers from across disciplines can move from a research question and data source to analysis and interpretation. An accompanying GitHub repository provides a ready-to-use implementation of the framework for existing and custom datasets.
\end{abstract}

\section{Introduction}
The study of `the mind' has undergone a notable change. Long confined to philosophical analysis or shielded from empirical inquiry as an unobservable and ostensibly unscientific object of study \parencite{skinner_science_1953}, mental information structures are now of explanatory interest \parencite{chomsky_review_1959}, not only in cognitive science but, and increasingly so, across the (social) sciences \parencite{kaidesoja_arguments_2019}.

Among these mental structures, concept representations form a core class. They emerge from individuals' experiences with the world \parencite{margolis_learning_2011} and encode the meanings individuals hold for concepts \parencite{kiefer_conceptual_2012}. As such, they serve as building blocks of thought, underpinning categorization, inference, memory, decision-making, action planning, and communication \parencite[]{margolis_concepts_2023}.

Evidence shows that the private meanings individuals hold for concepts often diverge markedly \parencite{hampton_investigating_2020}: two people may use the same word while understanding the concept it expresses in different ways. As the explanatory potential of private mental information structures is increasingly recognized, this raises theoretical questions beyond disciplinary boundaries. The present paper takes this challenge as its starting point, introduces concept representations and their analytical potential, and presents a measurement framework that enables concept representations to be modeled and interindividual variation in them to be analyzed.

Existing work has begun to use standardized textual traces to model concept representations through their representation in high-dimensional embedding spaces, both at the group level \parencite{aeschbach_mapping_2025} and at the individual level \parencite{petrenco_centroid_2026}. A key challenge, however, has been the stable modeling of the individual as the unit of analysis, partly due to the use of static word embeddings, which collapse variation across speakers and contexts into a single decontextualized vector per word \parencite{petrenco_centroid_2026}. Nor has this modeling been extended to the analysis of interindividual variation.

The present measurement framework addresses both gaps. Short textual responses as traces of participants' concept representations are elicited and encoded as transformer-based sentence embeddings, mapping entire utterances to points in a shared, high-dimensional vector space \parencite[]{reimers_sentence-bert_2019}, enabling stable representations even from short textual inputs \parencite{xu_revealing_2025}. For each concept, pairwise distances between participants' embeddings are then analyzed in a dyadic structure.

On this basis, I develop three analytical avenues, providing researchers with a typology of questions that can be addressed: (1) comparing concepts by how widely their representations are shared, (2) identifying variables that correlate with variation, and (3) examining how specific actors or subgroups deviate from a reference population in their concept representations. Each avenue is formalized and then illustrated in a proof-of-concept demonstration, guiding researchers through the full pipeline, from data source and research question to analysis and interpretation.

This paper builds on \textcite{durstewitz_concept_2026}, where we developed, validated, and demonstrated the measurement framework. In this paper, I extend the framework and make it available to researchers across disciplines. An open repository, available at \coderepo, allows researchers to apply it directly to their own concept sets, datasets, and predictor variables.

I begin by establishing the cross-disciplinary explanatory relevance of concept representations (Section \ref{sec:theoretical_motivation}). I then introduce the measurement framework and develop three analytical avenues (Section \ref{sec:measurement_framework}), each illustrated through a proof-of-concept demonstration (Sections~\ref{sec:present_study}--\ref{sec:demonstrations}).

\section{Theoretical Motivation}
\label{sec:theoretical_motivation}
The motivation for this paper rests on three observations. (1) Mental information structures are of increasing explanatory interest across disciplinary boundaries. (2) Concept representations form a core class of such structures. (3) Because the meanings they encode vary across individuals, this variation becomes an inherently interdisciplinary question.

\subsection{Mental Information Structures Across Disciplines}

Explanatory interest in mental information structures has expanded far beyond cognitive science. While finding early traction as a challenge to behaviorism in linguistics \parencite{chomsky_language_1968} and psychology \parencite{neisser_cognitive_1967}, a cognitive turn saw numerous disciplines shift from treating the individual as a black box toward examining the internal mental processes underlying human behavior \parencite{chomsky_review_1959}.

In sociology, for example, it challenged accounts in which culture is understood as an external sui generis force that programs individuals uniformly during socialization \parencite{dimaggio_culture_1997}. Instead, alongside public culture---external symbolic systems such as media messages, material artefacts, and institutionalized ``scripts''---personal culture is recognized as another level of analysis consisting of internalized schemas and associations encoded in the brain \parencite{cerulo_rethinking_2021,boutyline_cultural_2021} that may differ across individuals \parencite{lizardo_improving_2017}.

In economics, the principle of bounded rationality, according to which actors act on the basis of internal models of a situation rather than the situation itself \parencite{simon_behavioral_1955}, paved the way for behavioral economics \parencite{kahneman_prospect_1979}, shifting the explanatory focus from axiomatic models of rational choice toward the mental models and cognitive constraints that shape economic decision-making.

Comparable approaches can further be found, among others, in political science \parencite{nullmeier_cognitive_2010}, international relations \parencite{jervis_perception_1976}, history \parencite{duby_histoire_1961}, and geography \parencite[]{lynch_image_1960, mark_cognitive_1999}. While these approaches differ in their specific instantiations, they converge in recognizing mental structures of some sort as a private, mediating layer between the external world and behavior.

\subsection{Concept Representations}
While philosophical analysis has often treated concepts as abstract, mind-independent entities existing outside space, time, and psychology \parencite{frege_gedanke_1918}, or as having their extension fixed by the external world \parencite{putnam_meaning_1975}, they are today widely understood, at least within the empirical sciences, as learned mental representations and a core component of mental information structures \parencite{margolis_ontology_2007, margolis_learning_2011}.

Accordingly, even if concepts exist as mind-independent entities, individuals are assumed to associate a private understanding of them that may be partial or mistaken, while still shaping inference and behavior \parencite{burge_concepts_1993, burge_individualism_1979}. The present study terms such private understandings \textit{concept representations}: information structures capturing the meanings individuals hold for categories, objects, and ideas \parencite{kiefer_conceptual_2012} and through which they organize and interact with reality \parencite{von_glasersfeld_radical_1995}.

\subsection{Interindividual Variation in Concept Representations}
Empirical research has shown that the meanings individuals ascribe to concepts differ markedly \parencite{hampton_investigating_2020}. This concerns both what defines a concept \parencite{bellezza_reliability_1984-1,levitis_behavioural_2009} and which entities fall under it \parencite{mccloskey_natural_1978,bellezza_reliability_1984,barsalou_instability_1987}. Notably, individuals are largely unaware of this variation \parencite{marti_latent_2023}, and researchers have similarly left this variation ``underexplored'' \parencite[p.~5]{visconti_di_oleggio_castello_individual_2025}.

This neglect is noteworthy, given that concept representations are among the core mental information structures whose explanatory relevance now spans disciplinary boundaries. Studying them promises to surface latent differences that carry profound consequences, given their relevance to processes that concern both the individual and social life more broadly. Such differences may manifest in how people think and navigate the world, for example how they categorize, draw inferences, and learn \parencite[e.g.,][]{djalal_relationship_2018}. Furthermore, representational variation could explain how misunderstandings arise in communicative settings, potentially offering an underappreciated explanation for failures of mutual understanding and disagreement \parencite{marti_latent_2023}.

To model individuals' concept representations and analyze variation in them, I present in what follows a measurement framework. I first motivate its methodological components and then turn to three analytical avenues, offering a typology of questions the framework allows researchers to address.

\section{The Measurement Framework}
\label{sec:measurement_framework}
The framework elicits short textual responses as linguistic traces of concept representations, encodes these in embedding space, and compares vector representations geometrically. The following sections describe and motivate each step in turn.

\subsection{Textual Elicitation}

\begin{figure*}[t]
\centering
\includegraphics[width=.9\linewidth]{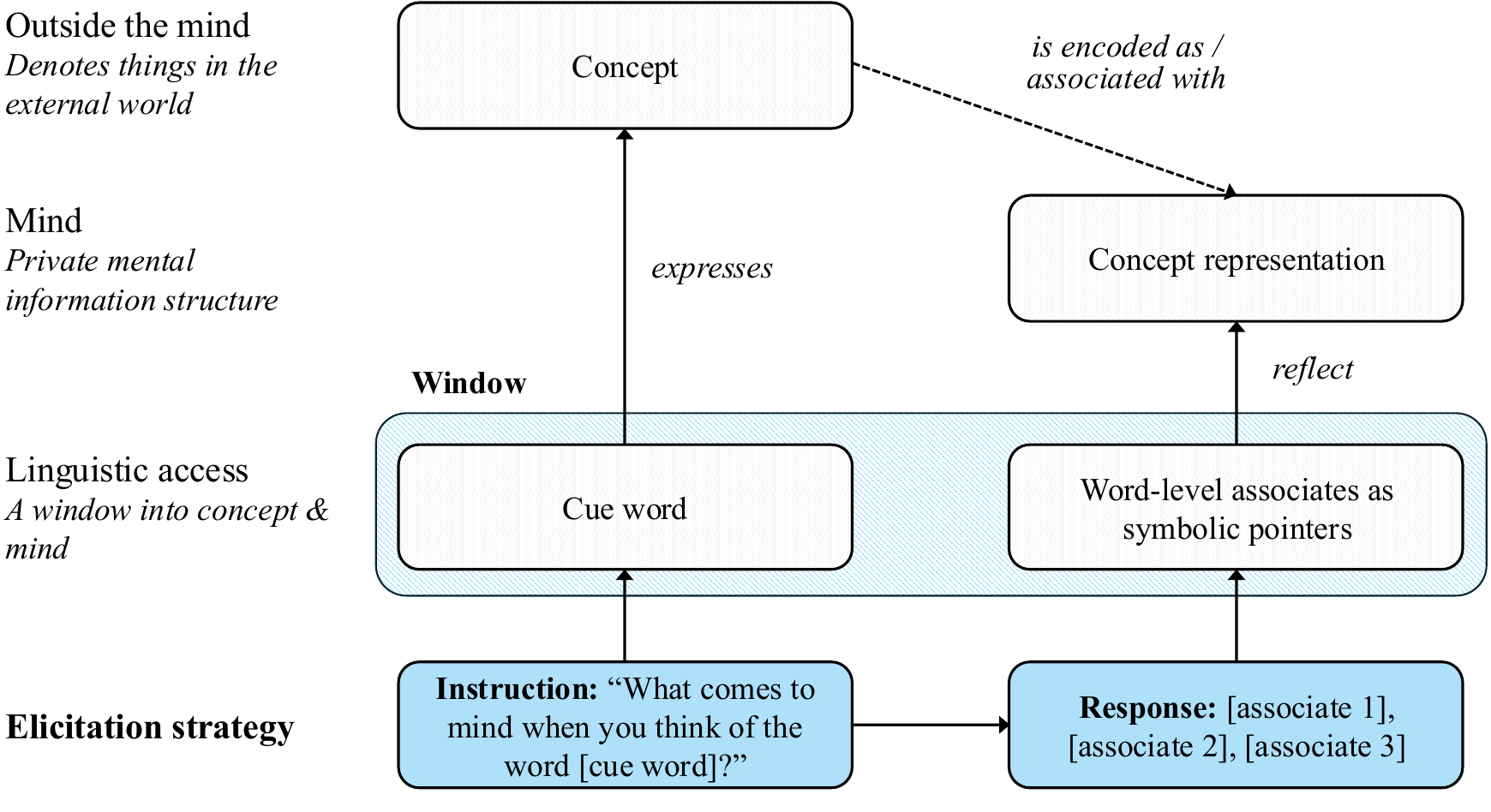}
\caption{\textbf{Free Association as One Elicitation Strategy.} The demonstrations in the present study use free association from the Small World of Words dataset \parencite[]{de_deyne_small_2019}. A cue word is assumed to express a concept, eliciting associates that reflect content salient to the participants' representation of that concept and are collectively assumed to reflect the underlying representation.}
\label{fig:premise}
\end{figure*}

The framework elicits short textual traces from participants in response to a concept under standardized conditions (Figure~\ref{fig:premise}).
In the instruction, a target word expresses a concept \parencite{frege_uber_1892}, for which a concept representation is then elicited. While words are often polysemous and can map onto multiple concepts, the target word should be disambiguated such that participants understand it as expressing the same concept.

Participants then articulate aspects of their concept representation in an open-ended response field. This approach draws on a long tradition that understands language as a “mirror of mind” \parencite[von Humboldt, as cited in][p.~67]{chomsky_language_2006}. Although language and mental representations constitute distinct levels of analysis \parencite{malt_representing_2024}, language is argued to systematically relate to \parencite{fodor_language_1975} and provide privileged access to mental representations, with words functioning as signs of underlying representations \parencite{ogden_meaning_1923, von_glasersfeld_radical_1995}.

The open-ended format allows participants to articulate their representations without constraining responses to the researcher’s prior assumptions, enabling a bottom-up investigation \parencite{petrenco_centroid_2026, aeschbach_mapping_2025}. Standardization holds context, purpose, and register constant, reducing confounds in naturalistic text data and making observed variation more attributable to representational differences. Various elicitation formats are compatible with the framework and are presented in Section~\ref{sec:data_method}, including their advantages and limitations.

\subsection{Encoding Responses as Sentence Embeddings}
The elicited responses are then mapped onto a fixed-length vector in a shared embedding space by a pre-trained transformer-based sentence encoder \parencite{reimers_sentence-bert_2019}. Through self-attention, these encoders weight each token in relation to the others, producing contextualized representations that capture the compositional structure of the full expression and disambiguate words by context \parencite{vaswani_attention_2017}.

Embedding spaces are argued to preserve semantic structure through statistical regularities learned from large text corpora \parencite{bengio_neural_2003}, operationalizing the hypothesis that semantic structure can be inferred from word–context co-occurrence patterns \parencite{firth_synopsis_1957, harris_distributional_1954}. While individual dimensions are not directly interpretable, directions between vectors carry semantic information \parencite{mikolov_efficient_2013, mikolov_linguistic_2013}. Sentence encoders are further fine-tuned on semantic similarity judgments, placing semantically similar expressions closer together in the embedding space. This motivates using geometric distance between embeddings as a measure of semantic dissimilarity. In embedding spaces, cosine similarity is the standard metric for this purpose, as Euclidean distance is unreliable in high-dimensional spaces \parencite{aggarwal_surprising_2001}.

\subsection{Dyadic Modeling of Pairwise Distances}
\begin{figure*}[p]
\centering
\includegraphics[width=1\linewidth]{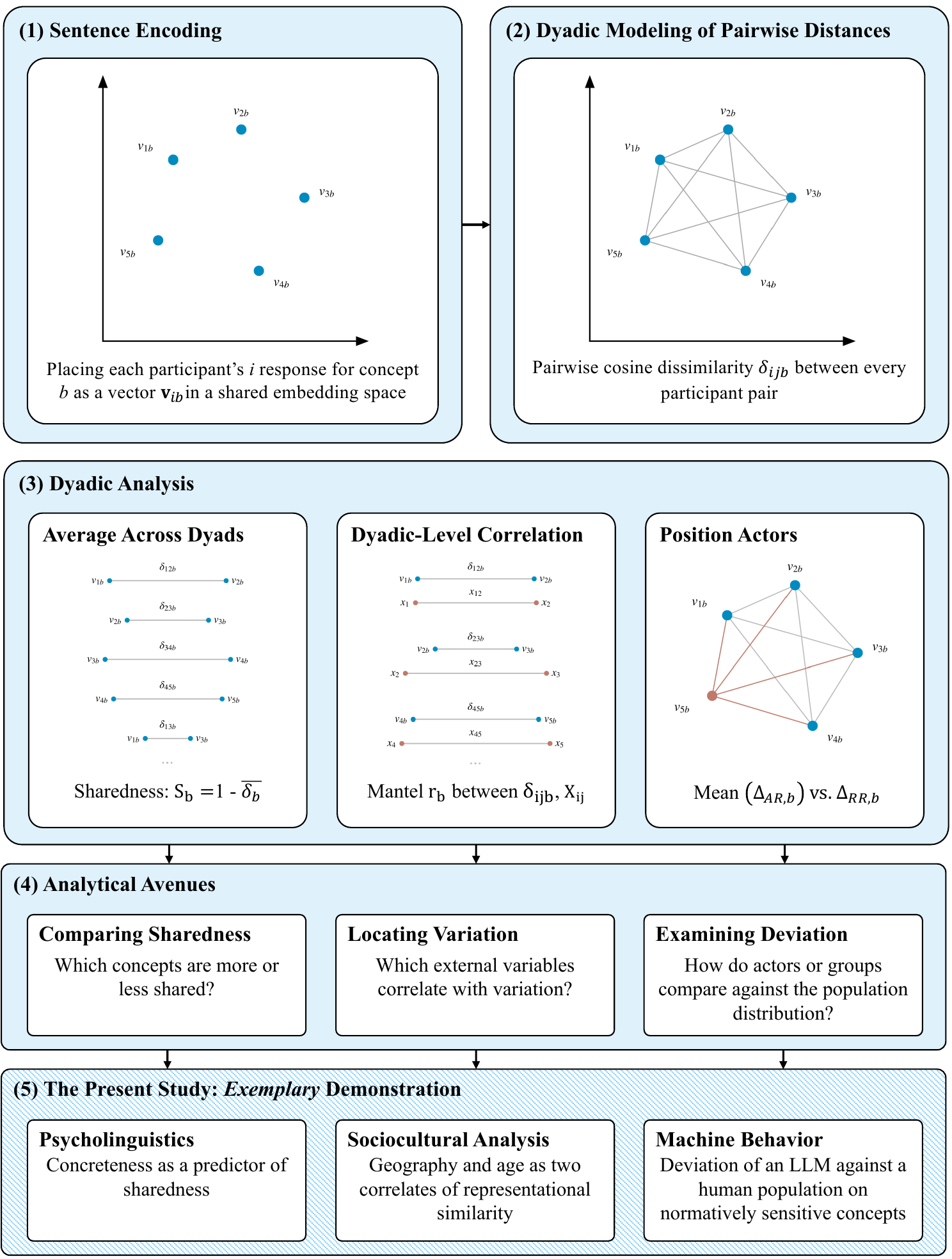}
\caption{\textbf{From Pairwise Distances to Three Analytical Avenues.} Textual responses are (1) encoded as sentence embeddings $\mathbf{v}_{ib}$, and (2) pairwise cosine dissimilarities $\delta_{ijb}$ form the point of departure for analyzing representational variation. Panel (3) illustrates three operations the dyadic statistical modeling enables. Panels (4) and (5) map each operation to its analytical avenue and exemplary demonstration.}
\label{fig:framework}
\end{figure*}

Variation is analyzed, for each concept, by comparing each participant's embedding with every other participant's. For a given concept $b$, the cosine dissimilarity between participants $i$ and $j$ is computed from their unit-normalized embeddings as:
\begin{equation}
    \delta_{ijb} = \frac{1 - \mathbf{v}_{ib}^\top \mathbf{v}_{jb}}{2} \in [0, 1]
\end{equation}
where $\delta_{ijb}$ measures representational distance between participants $i$ and $j$ on concept $b$; lower values indicate greater similarity and higher values greater divergence. This yields $N(N-1)/2$ unordered participant pairs per concept in a round-robin dyadic structure \parencite[]{kenny_social_1984}. The resulting pairwise dissimilarities $\delta_{ijb}$ form the basis for the three analytical avenues developed in the following sections and illustrated in Figure~\ref{fig:framework}.

\subsubsection{Comparing Sharedness.}
\label{sec:comparing-sharedness}

Research has long pointed to interindividual differences in concept representations \parencite{hampton_investigating_2020}. A pertinent question is whether some concepts exhibit greater representational sharedness than others. From an applied perspective, researchers may be interested in identifying which concepts of disciplinary relevance are relatively shared or fragile in their representational structure. From a more foundational perspective, concepts differ systematically in properties such as conceptual domain \parencite{atran_basic_1989}, taxonomic level \parencite{rosch_principles_1978}, or degree of concreteness \parencite{brysbaert_concreteness_2014}, raising the question of whether representational sharedness varies systematically along these dimensions.

The dyadic structure of the framework makes concepts comparable by constructing a continuous estimand of representational sharedness. For each  concept $b$, the mean pairwise representational dissimilarity across all participant dyads is computed as:
\begin{equation}
    \bar{\delta}_b = \frac{2}{N(N-1)} \sum_{i < j} \delta_{ijb}
\end{equation}
Sharedness ($S_b$) is then defined as its complement:
\begin{equation}
    S_b = 1 - \bar{\delta}_b
\end{equation}
such that high $S_b$ indicates greater representational similarity across 
individuals. To illustrate the range of analytical questions opened by such analyses, consider the following: 
\begin{itemize}
\setlength\itemsep{1pt}
\setlength\parskip{0pt}
    \item Do ``thick'' ethical concepts (e.g., \textit{cruel}, \textit{brave}, 
    \textit{dishonest}) exhibit greater sharedness than ``thin'' ones 
    (e.g., \textit{good}, \textit{right}, \textit{wrong}) 
    \parencite{williams_ethics_1985}?
    \item Are concepts acquired earlier in development represented more 
    similarly across individuals?
    \item Are natural kind concepts more shared than social or artefactual 
    concepts?
\end{itemize}

\subsubsection{Locating Variation.}

Given a concept's degree of sharedness, a complementary question concerns how its variation is structured. Individuals are argued to encode concept representations through experiential encounters \parencite{margolis_learning_2011}, including direct sensorimotor interaction with concept referents and linguistically mediated forms of acquisition such as books, formal instruction, and media consumption. These experiential histories are themselves structured by the social, cultural, and environmental contexts in which individuals are embedded \parencite{burge_individualism_1979, berger_social_1967, kramsch_language_2014}. It is therefore plausible that differences in concept representations systematically correlate with variables reflecting differential forms of exposure and experience.

The framework allows researchers to examine such variables by comparing, for each participant dyad $(i, j)$, representational distance $\delta_{ijb}$ against a corresponding predictor distance. For a participant-level predictor $p$, the pairwise predictor distance is defined as:
\begin{equation}
    x_{ij} = |p_i - p_j|
\end{equation}
A Mantel test then assesses the significance of the correlation $r_b$ between $\delta_{ijb}$ and $x_{ij}$ across all dyads for concept $b$. To illustrate the range of analytical questions opened by such analyses, consider the following:
\begin{itemize}
\setlength\itemsep{1pt}
\setlength\parskip{0pt}
\item Do differences in cultural capital correlate with representational similarity for concepts marking aesthetic distinction \parencite{bourdieu_distinction_1984}?
\item Does informational diet correlate with representational similarity for concepts central to public debate?
\item Does childhood urban density correlate with representational similarity for concepts of community and public space?
\end{itemize}

\subsubsection{Examining Deviation.}
\label{sec:deviation}

Researchers may be interested not only in how shared concepts are on average, but also in the extent to which specific actors or groups deviate from the broader population in their representation of particular concepts. For some groups, such as clinical populations, domain experts, or cultural communities, systematic deviations from the broader population may be theoretically expected or socially consequential.

The framework supports such analyses by extending the round-robin structure. Rather than comparing every participant against every other participant, a focal actor or subgroup can be compared against a reference population.

For a focal actor or subgroup $A$ and a reference population $R$, two pairwise cosine dissimilarity distributions can be constructed for each concept $b$. The within-reference distribution,
\begin{equation}
\Delta_{RR,b} = \{\delta_{ijb} \mid i,j \in R,\ i < j\},
\end{equation}
captures representational variability within the reference population itself. The cross-group distribution,
\begin{equation}
\Delta_{AR,b} = \{\delta_{ijb} \mid i \in A,\ j \in R\},
\end{equation}
captures the representational distance of the focal actor or subgroup to every participant in the reference group (see also Figure~\ref{fig:framework}). Comparing the two distributions therefore locates the focal actor or subgroup relative to typical interindividual variation within the reference population. Statistical significance can be assessed by comparing the observed mean of $\Delta_{AR,b}$ against a permutation distribution sampled from $\Delta_{RR,b}$. \footnote{This extends naturally to multi-group comparisons, though this case is not developed here given existing work \parencite{aeschbach_mapping_2025}.}

To illustrate the range of analytical questions opened by such analyses, consider the following:

\begin{itemize}
\setlength\itemsep{1pt}
\setlength\parskip{0pt}

    \item Do individuals with psychopathy deviate from the broader population in their representations of concepts such as empathy, harm, or remorse?
    
    \item Do judges hold representations of concepts such as justice, fairness, or punishment that systematically deviate from those of the populations they govern?
    
    \item Do trained economists represent concepts such as fairness, 
    efficiency, and welfare differently from a lay reference population?

\end{itemize}

\section{The Present Study}
\label{sec:present_study}

The present study illustrates each analytical avenue through a proof-of-concept demonstration, guiding researchers from the choice of data source and research question through analysis and interpretation.

\textit{Comparing sharedness} is examined through a psycholinguistic demonstration asking whether concepts are more shared at higher levels of concreteness. \textit{Locating variation} is illustrated by a sociocultural demonstration examining whether absolute latitude, as a proxy for differential ecological exposure, correlates with representational similarity for a set of climate and ecology concepts, and whether age, as a proxy for generational cultural exposure, does so for a set of sociopolitical and economic concepts. \textit{Examining deviation} is addressed through a machine behavior demonstration that treats an LLM as a social actor and assesses its representational alignment with a human population on normatively sensitive concepts.

\section{Data and Method}
\label{sec:data_method}
The measurement framework operates on short textual responses elicited under standardized conditions. To access concept representations, several elicitation formats have been used, including full-response formats such as participant-generated definitions \parencite{bellezza_reliability_1984-1}, and word-level formats such as feature listing \parencite{bellezza_reliability_1984, bhatia_distributed_2019}, word substitution \parencite{pugacheva_lexical_2024}, and free association \parencite{aeschbach_mapping_2025}. These formats involve different trade-offs.

Full-response formats such as elicited definitions, or potentially other extended responses such as explanations or open-ended reflections, let participants draw on the compositionality and relationality of language, expressing how a concept's properties relate to one another and to other concepts. This can help make complex aspects of a concept's representation explicit \parencite{jackendoff_languages_1992}, offering semantic \textit{depth}. Such data are, however, rarely available at scale and require custom collection. While the framework can operate on relatively small samples \parencite{durstewitz_concept_2026}, collecting such data may nevertheless involve substantial economic costs and logistical effort.

Word-level response formats, by contrast, lack inherent compositional structure but offer considerable \textit{breadth} through the availability of large-scale data sources comprising thousands of cue words. Existing datasets include the \textit{Small World of Words} (SWOW) \parencite{de_deyne_small_2019} and the \textit{University of South Florida Free Association Norms} \parencite{nelson_university_2004} for free association, as well as the \textit{English Semantic Feature Production Norms} \parencite{buchanan_english_2019} for feature listing.

The present demonstrations draw on the SWOW dataset. A series of preprocessing steps transforms word-level association responses into structured sentence-level inputs that can leverage the contextual capabilities of transformer-based encoders, illustrating how existing large-scale data sources can be incorporated into the framework. The following sections introduce free association and the SWOW dataset before describing preprocessing and encoding.

\subsection{Data Source}

Free association has a long history in psychology and is argued to reveal properties of both the processes and representations underlying semantic cognition \parencite{de_deyne_small_2019}. Presented with a cue word expressing a concept \parencite[]{frege_uber_1892}, participants produce the first words that come to mind \parencite{nelson_what_2000}, generating word-level pointers to related concepts, properties, examples, and other content salient to their representation of the concept.

The SWOW English word association norms \parencite[]{de_deyne_small_2019} is a large-scale open-source dataset collected between 2011 and 2018. Participants were recruited online and asked to respond to each cue word with the first three words that came to mind, yielding up to three responses per cue (R1, R2, R3). The dataset comprises responses from 83,864 participants across 12,218 cue words, resulting in 3,684,600 responses in total. Alongside the association data, each participant record includes self-reported sociodemographic information covering age, native language, education level, gender, and city and country of residence. Participants were predominantly native English speakers (81\%) with a  mean age of 36 years ($SD = 16$). 

\subsection{Data Transformation}

A main limitation of the data is that associates as isolated tokens yield static, decontextualized representations in which polysemy is unresolved and embeddings reflect distributional averages over all usage contexts \parencite[e.g.,][]{turney_frequency_2010, peters_dissecting_2018}, making individual-level representation unstable. When presented as an unstructured sequence to a transformer-based encoder, associates are processed as a token list with weak positional and syntactic constraints, leaving relations among associates, and between each associate and the cue word, unspecified \parencite[][]{wang_language_2019, vaswani_attention_2017}.

To address this, responses were preprocessed and transformed as follows. First, associates were cleaned by stripping punctuation and whitespace, and responses consisting of SWOW placeholders (``No more responses'', ``Unknown word'') were excluded. Second, only responses with three valid associates were retained to ensure comparability across participants and sufficient semantic content for sentence-level encoding. Third, the cue vocabulary was restricted to (1) nouns, identified via WordNet part-of-speech lookup \parencite{miller_wordnet_1995}, reflecting the view that concepts are paradigmatically denoted by nouns \parencite[][p.~194]{gardenfors_conceptual_2004}, and (2) cue words with at least $N = 75$ responses.

Finally, these three associates were transformed into \textit{synthetic sentences} according to a uniform template: ``[Cue word] is associated with [R1], [R2], and [R3].'' The sentence template provides a syntactic scaffold that activates the attention mechanism in transformer-based encoders, weighting each associate jointly in relation to the others and to the cue word, producing a contextualized, compositionally structured representation of the participant's response.

\subsection{Encoding}
Synthetic sentences were encoded using \texttt{all-\allowbreak mpnet-\allowbreak base-\allowbreak v2}, a state-of-the-art general-purpose sentence encoder with strong benchmark performance \parencite{song_mpnet_2020}. Where resources permit, researchers can alternatively draw on larger encoder models within the SBERT framework \parencite[]{reimers_sentence-bert_2019}, or on LLMs with decoder-only architectures adapted as text encoders, which have been shown to achieve higher benchmark performance than many SBERT-family models \parencite{wang_improving_2024}.

\section{Demonstrations}
\label{sec:demonstrations}

With the preprocessed and transformed responses prepared as the data source, the following sections walk through the three demonstrations. All three are intended as proofs of concept rather than as substantive empirical findings. General limitations are discussed subsequently.

\subsection{\textit{Comparing Sharedness}: Concreteness as a Predictor of Sharedness}

Concreteness is a psycholinguistic variable referring to the degree to which a concept expressed by a word refers to a perceptible entity \parencite{brysbaert_concreteness_2014, solovyev_concretenessabstractness_2021}. Along this continuum, concrete concepts refer to material objects with direct referents in the physical world, whereas abstract concepts refer to mental states, conditions, or relations.

Concrete concepts may give rise to more shared representations because they are anchored in a common external referent \parencite{hacking_tradition_1991, cohnitz_variation_2020}, whereas abstract concepts, encountered through varying linguistic experience \parencite{lenci_emotions_2018}, may permit greater variability \parencite{johns_determining_2024}. This gradient finds empirical support in existing research, with interindividual variation higher for abstract than concrete concepts in definition tasks \parencite{bellezza_reliability_1984-1}, large-scale naturalistic data \parencite{johns_determining_2024}, and with fMRI response variability higher in regions associated with abstract and social processing than in those associated with physical objects \parencite{visconti_di_oleggio_castello_individual_2025, wang_fmri_2022}.

The measurement framework, combined with large-scale free-association and concreteness norms, allows this hypothesis to be examined at unprecedented scale.

\begin{figure*}[t]
\centering
\includegraphics[width=0.7\textwidth]{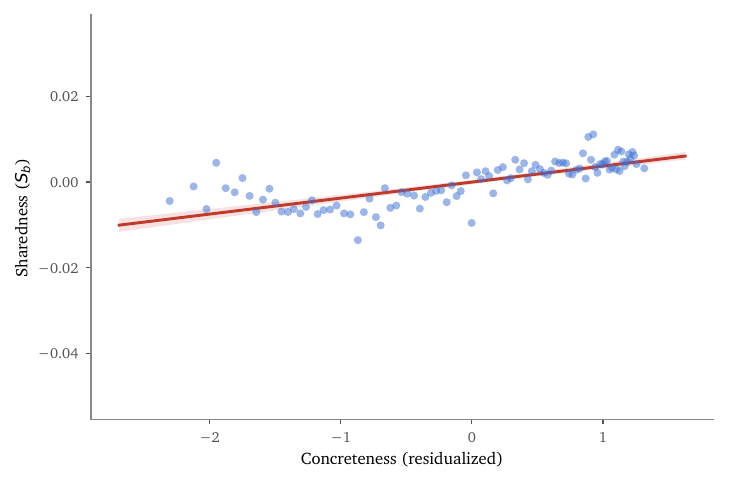}
\caption{\textbf{Sharedness as a Function of Word Concreteness, Controlling for WordNet Synset Count.} Each point represents the mean $S_b$ of concepts falling within one of 100 equal-frequency bins along the residualized concreteness axis, used here for visual clarity. The red line shows the weighted regression of the residualized outcome on residualized concreteness, with 95\% bootstrap confidence intervals derived from 2{,}000 resampled iterations.}
\label{fig:concreteness_regression}
\end{figure*}

\subsubsection{Method.}
Concreteness ratings were drawn from \textcite{brysbaert_concreteness_2014}, a large-scale psycholinguistic norm database providing mean concreteness scores on a 1--5 scale for 37,058 English words, collected from over 4,000 crowd-sourced raters. The SWOW cue vocabulary was matched against this database case-insensitively, retaining only cues for which a rating was available. This yielded 7,619 concepts with $N = 721{,}353$ responses in total.

Sharedness ($S_b$) was computed per concept and the relationship between concreteness and $S_b$ assessed as a partial correlation. A potential confound arises from polysemy. Because SWOW cue words are presented without disambiguating context, participants may have produced associations to different concepts. If polysemy co-varies with concreteness, differences in representational variability across the concreteness spectrum may partly reflect the former. To address this, WordNet synset count was included as a covariate, representing the number of distinct senses assigned to a word in WordNet and serving as a proxy for polysemy \parencite{miller_wordnet_1995}. 

The partial correlation was computed via residualization. Both the outcome and concreteness were regressed on synset count with observation counts $N$ as weights, and Pearson's $r$ between the resulting residuals is reported with 95\% bootstrap confidence intervals.

\subsubsection{Results.}
Figure~\ref{fig:concreteness_regression} presents the partial regression of $S_b$ on word concreteness, controlling for WordNet synset count. Concreteness was positively correlated with $S_b$ ($r = .166$, 95\% CI $[.146, .187]$, $p < .001$).

\subsubsection{Discussion.}
The demonstration replicates prior findings while extending evidence to a substantially larger concept set and free-association data. The observed correlation was modest, indicating that concreteness captures only part of the variation in $S_b$.

A  limitation concerns the comparability of $S_b$ across concepts. Because the comparison spans concepts from a wide range of semantic domains, and embedding spaces may not encode all domains at equal resolution \parencite{wang_unsupervised_2023}, differences in $S_b$ across concepts may partly reflect embedding space geometry. Future work may include general word frequency as a proxy covariate for non-uniform semantic resolution.

\subsection{\textit{Locating Variation}: Geographic Location and Age as Predictors of Representational Similarity}

Concept representations are shaped by individuals' experiential histories, encompassing both direct sensorimotor encounters with concept referents and linguistically mediated acquisition through cultural and historical exposure \parencite{margolis_learning_2011}. This demonstration examines two predictors of representational variation: absolute latitude as a proxy for ecological and climatic environment, and age as a proxy for generational cultural exposure.

Latitude may correlate with representational differences because different regions expose individuals to different physical environments. The representation of a concept such as \textit{tree} may, for example, be shaped by everyday exposure to broadleaf trees at lower latitudes or to conifers at higher latitudes. This expectation is consistent with research showing that geographic and cultural contexts are associated with systematic differences in vocabulary, semantic boundaries, and cognitive organization, particularly in domains such as plants, animals, and landscape features \parencite{leavitt_ethnosemantics_2014,bromhead_ethnogeographical_2011,brown_language_2014}.

Age may correlate with representational differences because different generations are exposed to different cultural and historical environments \parencite[e.g.,][]{ryder_cohort_1965}. Consistent with this possibility, research on semantic drift shows that aggregate word meanings shift systematically over time in large text corpora \parencite{hamilton_diachronic_2016,tahmasebi_survey_2019}, but leaves open whether such changes are reflected at the individual level.

\subsubsection{Method.}
Two theoretically motivated sets of 25 concepts were assembled, one for each predictor. A climate and ecology set was used to examine whether representational differences correlate with absolute latitude, reflecting differential exposure to physical and ecological environments. A sociopolitical and economic set was used to examine whether representational differences correlate with age, reflecting differential cultural and historical exposure.

Geographic coordinates were obtained by matching participants' normalized self-reported city names and ISO country codes to the GeoNames cities500 database \parencite{geonames_geonames_2026}, retaining the most populous entry for ambiguous matches. Based on these coordinates, absolute latitude ($p = |\lambda|$) served as one predictor. The other predictor was self-reported age ($p = a$) in years. Pairwise predictor distances $x_{ij} = |p_i - p_j|$ were then correlated with $\delta_{ijb}$ using one-tailed Mantel tests. Dyads with missing values and cue words with fewer than 75 valid dyads were excluded.

\subsubsection{Results.}

\begin{figure*}[t]
\centering
\includegraphics[width=1\linewidth]{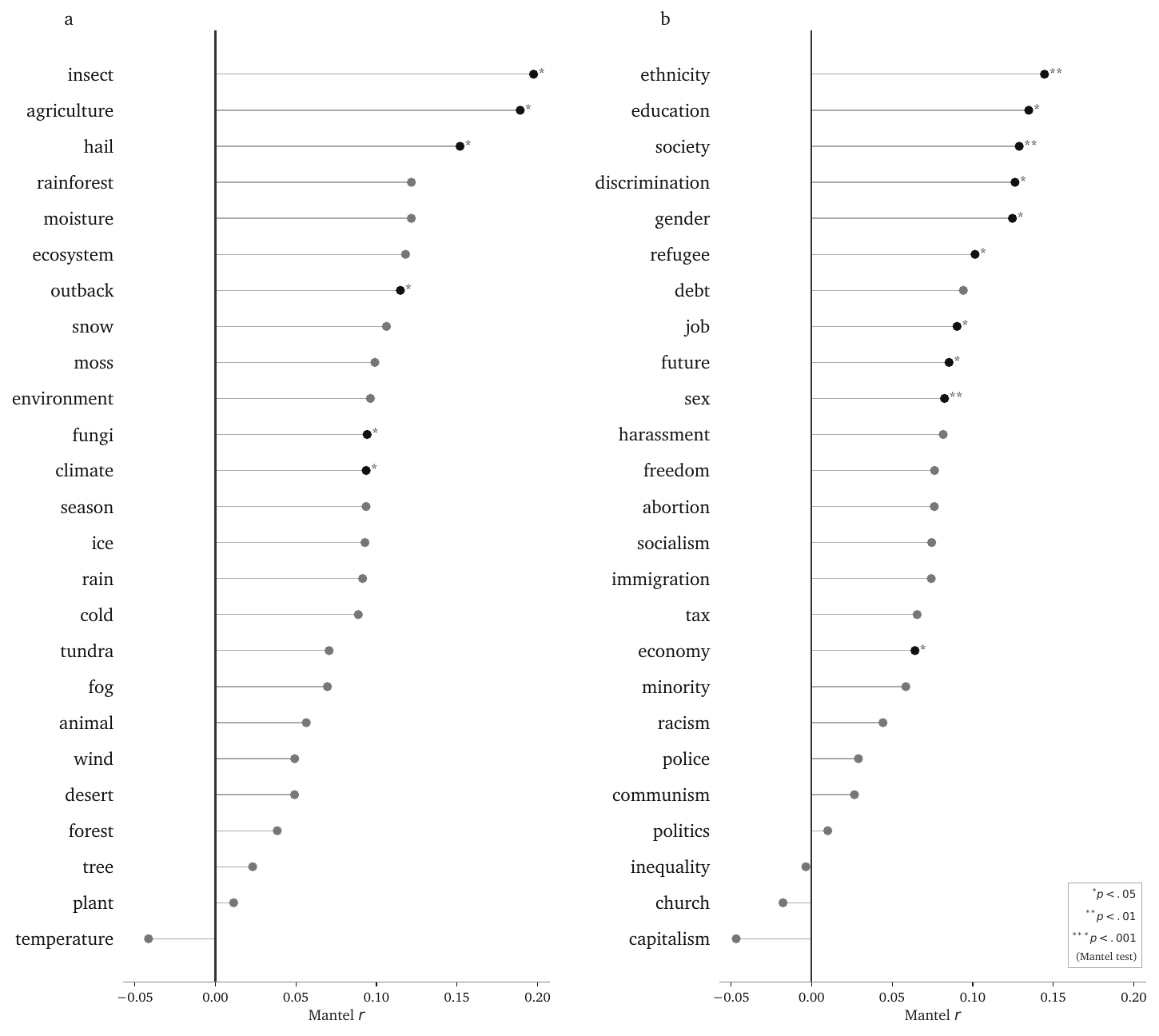}
\caption{\textbf{Concept-Level Correlations Between Representational Differences and Predictor Distances.} Panel (a) shows results for the 25 climate and ecology concepts analyzed against difference in absolute latitude; panel (b) shows results for the 25 sociopolitical and economic concepts analyzed against difference in age. Each point represents one concept. Filled points reach the significance threshold indicated by the asterisk markers. $r$ = Pearson correlation between $\delta_{ij}$ and $x_{ij}$, assessed via one-tailed Mantel test with 9,999 permutation iterations.}
\label{fig:mantel_combined}
\end{figure*}

Figure~\ref{fig:mantel_combined} shows Mantel correlations for the two concept sets. Both predictors yield positive correlations for a subset of concepts, consistent with the directional predictions. For latitude, 6 of 25 concepts reached significance, with the strongest correlations observed for \textit{insect}, \textit{agriculture}, and \textit{hail}. For age, 10 of 25 concepts reached significance, with the largest correlations for \textit{ethnicity}, \textit{education}, and \textit{society}. Correlations showed substantial heterogeneity across concepts, reaching approximately $.20$ for some concepts, while others exhibited values near zero or in the negative direction. 

\subsubsection{Discussion.}
The demonstration illustrates how external variables such as latitude and age can be tested as predictors of differences in concept representations. The observed correlations were low to modest, however, reaching approximately $.20$ at most, and varied substantially across concepts, plausibly reflecting the many sources of variation beyond the predictors examined here. 

Several factors warrant caution. The concept sets were small, and the predictors themselves are coarse. Latitude, for instance, does not distinguish between locations at the same latitude that differ in elevation, urbanization, or proximity to coastlines, while age conflates differences in formative experience with more general age-related effects \parencite[e.g.,][]{peterson_people_2020,bleidorn_personality_2022}. Future work may draw on larger concept sets and more precise predictors of environmental and cultural exposure.

\subsection{\textit{Examining Deviation}: Representational Deviation of an LLM on Normatively Critical Concepts}

LLMs have become fixtures of everyday life, mediating social, cultural, economic, and political interactions \parencite{rahwan_machine_2019}. As their societal role expands, so do concerns about their consequences \parencite{weidinger_ethical_2021,bender_dangers_2021}.

Prior research has shown that LLMs can reproduce algorithmic bias and stereotyping by reinforcing harmful associations between social groups and attributes \parencite{bolukbasi_man_2016}, generate toxic or hateful content \parencite{gehman_realtoxicityprompts_2020}, and encode implicit normative assumptions without transparency \parencite{santurkar_whose_2023}. Whether LLM representations of normatively critical concepts align with human representations is therefore an open and normatively relevant question.

Because LLMs operate as black boxes that resist direct inspection, the interdisciplinary field of machine behavior proposes treating them as actors that can be examined empirically through their behavior \parencite{rahwan_machine_2019}. In line with this perspective, a growing body of research has investigated LLMs using behavioral experimental paradigms \parencite{hagendorff_machine_2024}. In this demonstration, an LLM is treated as a participant in a standardized elicitation protocol, allowing its responses to be encoded in the same embedding space as the human SWOW data and its representational alignment to be directly assessed.

\subsubsection{Method.}

A set of 25 normatively sensitive concepts, for which divergence between human and LLM representations may be particularly consequential, was assembled, spanning interpersonal violence and harm, identity-based discrimination, mental health, and properties relevant to the deployment of sociotechnical systems.

Associations were elicited from Gemini 2.5 Flash, a multimodal LLM developed by Google DeepMind and accessed through the Google AI API. The model was selected as a representative case of LLMs encountered by users in everyday deployment. The elicitation procedure replicated the SWOW protocol within an LLM prompting setup. The model received a system prompt framing the task (``You are participating in a word association study. Respond only with the requested words, nothing else.'') and a user prompt presenting the cue and the required output format (``Respond with the first three words that come to mind when you see the word \{cue word\}. Output format: word1, word2, word3. Nothing else.''). The elicitation was repeated 20 times per concept at a sampling temperature of 0.7, with each run producing a three-word response, yielding 20 responses per concept. Non-zero sampling temperature produces a distribution of responses across repeated runs of the same prompt, which the repeated elicitation procedure samples.

Responses were transformed into synthetic sentences using the identical template applied to the human data (``[Cue word] is associated with [R1], [R2], and [R3].'') and encoded with \texttt{all-mpnet-base-v2}.

For each concept, the LLM-human dissimilarity distribution $\Delta_{AR,b}$ was compared against the human-human reference distribution $\Delta_{RR,b}$, and significance assessed using a two-sided permutation test.

\subsubsection{Results.}

\begin{figure*}[p]
\centering
\includegraphics[width=1\linewidth, height=\textheight, keepaspectratio]{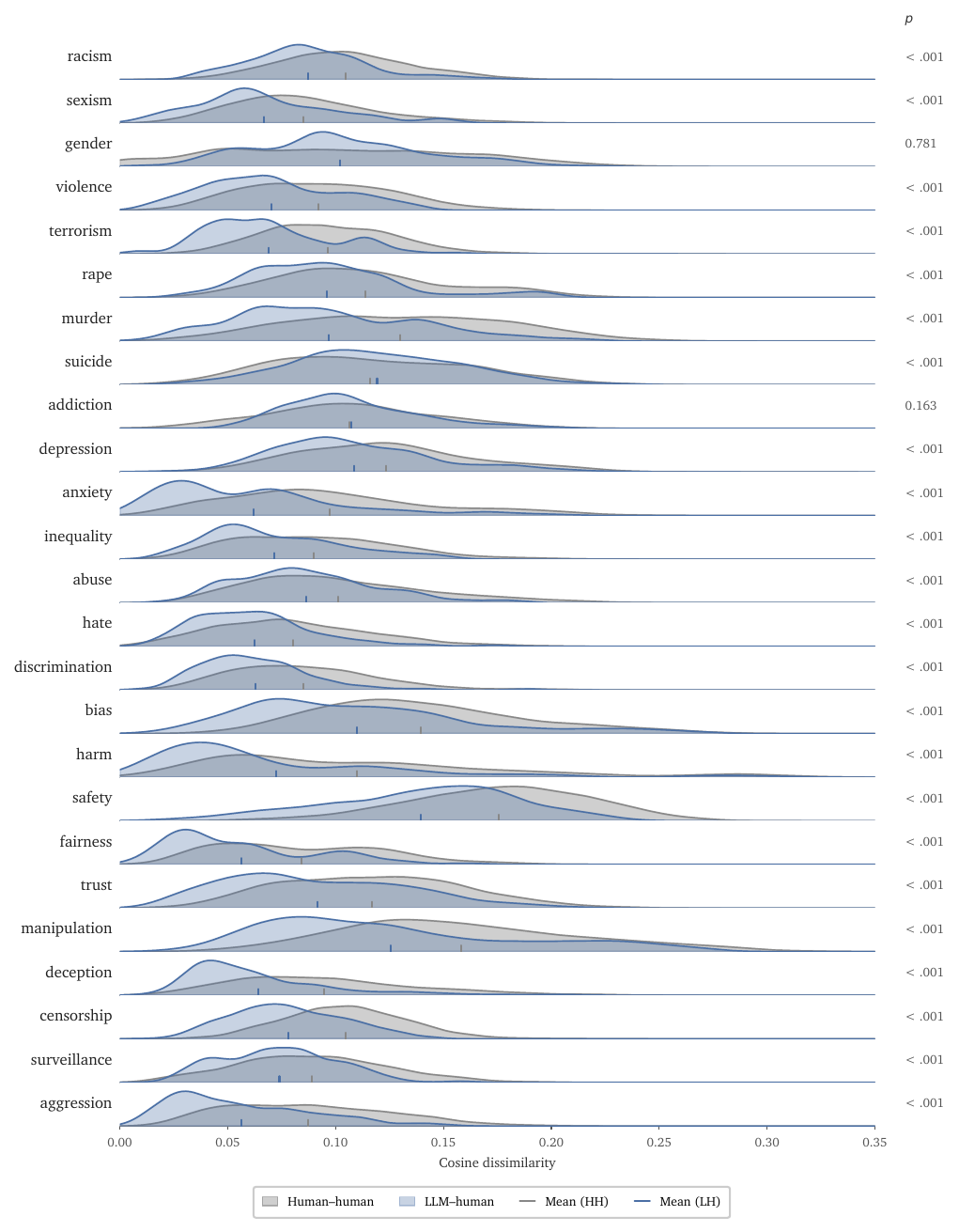}
\caption{\textbf{Pairwise Cosine Dissimilarity Distributions for Human--Human and LLM--Human Response Pairs.} Each panel shows the kernel density estimate of pairwise cosine dissimilarities for human--human pairs (grey) and LLM--human pairs (blue). Vertical tick marks at the base of each panel indicate the mean pairwise dissimilarity for each distribution. $p$-values are from two-sided permutation tests ($9{,}999$ iterations) assessing whether the mean LLM--human dissimilarity differs from the human--human baseline.}
\label{fig:alignment_distributions}
\end{figure*}

Figure~\ref{fig:alignment_distributions} shows kernel density estimates of pairwise cosine dissimilarities for LLM--human and human--human pairs across the 25 concepts. For 23 of the 25 concepts, the LLM--human distribution deviated from the human--human distribution (all $p < .001$), with gender ($p = .781$) and addiction ($p = .163$) as exceptions.

Across 22 of those 23 concepts, the LLM--human distribution was shifted left of the human--human distribution, with the LLM on average more similar to any given human than humans were to each other. Suicide was the sole reversal, with the LLM--human distribution shifted right.

An additional test addressed a separate concern. While the LLM may have learned a representational core shared by most humans, it may produce homogeneous responses clustered near the human average without reproducing human representational diversity. The analysis compared the variability of within-LLM and human--human pairwise dissimilarities. Variability was lower for the LLM across all 25 concepts (all $p < .001$). On average, human variability ($M_\text{SD} = .042$) was approximately 3.4 times greater than LLM variability ($M_\text{SD} = .012$).

\subsubsection{Discussion.}

The demonstration examined an LLM as a behavioral participant and assessed its deviation from a human population on a set of normatively sensitive concepts. Although the model was typically closer to individual humans than humans were to one another, an additional analysis revealed substantially lower representational variability across all concepts. One plausible explanation is that reinforcement learning from human feedback (RLHF) trains models toward averaged human preferences, potentially pulling outputs toward the population center and compressing variability \parencite{chaudhari_rlhf_2025}. 

Several limitations warrant caution. The analysis relied on a single model and a small, theoretically assembled concept set, and results may depend on the specific elicitation prompts used. Future work may extend the analysis to broader concept sets, alternative prompt formulations, and additional models.

\section{General Discussion}
\label{sec:discussion}
Concept representations encode the meanings individuals hold for categories, objects, and ideas \parencite{kiefer_conceptual_2012} and organize how they interpret and navigate the world \parencite{von_glasersfeld_radical_1995}. The present study presented a measurement framework for modeling individual concept representations and developed three analytical avenues for the empirical study of their variation: \textit{comparing sharedness}, \textit{locating variation}, and \textit{examining deviation}. Three demonstrations suggest that the framework can generate tractable insights across diverse substantive domains. Several challenges, however, warrant consideration.

\subsection{Elicitation Format}

The demonstrations draw on SWOW free-association data, whose breadth permits the analysis of thousands of concepts across a large participant population. Associates were transformed into synthetic sentences before sentence embedding to provide a standardized scaffold for transformer-based encoding. While free association constitutes a well-established elicitation technique, several limitations warrant consideration.

First, preprocessing can only leverage information present in the elicited responses. Although the synthetic sentence scaffold provides richer context for transformer-based encoding, the underlying free-association data remain sparse in compositional structure, limiting what participants can express about their concept representations, particularly complex aspects such as internal relational structure \parencite{jackendoff_languages_1992}. Second, free association presents cues without contextual disambiguation, such that responses from different participants may reflect distinct underlying concepts. Third, free association correlates only moderately with alternative measures, including self-reports and behavioral assessments \parencite{schnabel_free_2013,wulff_semantic_2022}.

Importantly, these limitations pertain to the elicitation format rather than the framework itself. The analytical pipeline can be applied to various elicitation strategies under standardized conditions. Researchers may select data sources according to their substantive questions and weigh the availability of existing resources against the costs of custom data collection. Future research may also triangulate across elicitation methods to distinguish method-specific findings from those that generalize across elicitation contexts.

\subsection{Concepts in a Vacuum?}

The measurement framework operates on textual traces elicited under standardized conditions, which helps circumvent the confounds of naturalistic language data. At the same time, these elicitation strategies, including the free association underlying the SWOW dataset, typically abstract from context, raising the question of whether concept representations can meaningfully be studied outside the situations in which they are used.

Doing so presupposes that concept representations have stable content that combines compositionally with other constituents \parencite{fodor_connectionism_1988}, an assumption in tension with contextualist accounts of meaning. One such prominent critique originates in Wittgenstein's later work, according to which meaning is constituted through use within ``language games'' rather than fixed semantic cores \parencite{wittgenstein_philosophical_1958}. On this view, concepts are always deployed within specific contexts and do not have stable meaning outside them. Eliciting a representation outside a concrete context therefore risks stripping away part of what gives it meaning.

Empirical research lends some support to this critique. Studies have shown that individuals assign objects to different concepts depending on their imagined use \parencite{labov_boundaries_1973}, that typicality judgments vary with situational framing \parencite{roth_effect_1983}, and that concept construction depends on social perspective \parencite{barsalou_constructing_1984}.

This suggests several directions for future research. First, the predictive validity of concept representations elicited under context-invariant conditions remains an open empirical question. An initial step in this direction was taken in earlier work, where we showed that individual concept representations predict typicality judgments \parencite{durstewitz_concept_2026}. Second, standardized elicitation need not exclude context altogether. Future research could systematically vary situational framing while holding the focal concept constant, providing a way to examine how concept representations vary across contexts of use.

\subsection{Sentence Encoding}

Sentence embeddings serve as measurement instruments for the study of concept representations. They make textual traces quantitatively comparable in vector spaces, where geometric proximity is taken to reflect semantic proximity. Sentence encoders produce contextualized representations, offer computational efficiency relative to alternative approaches such as distributed semantic models \parencite{johns_determining_2024}, and produce stable representations even from short textual inputs \parencite{xu_revealing_2025}. Several considerations nevertheless warrant discussion.

First, sentence encoders function as black-box measurement instruments. While representational differences can be quantified as cosine dissimilarities, the dimensions along which individual representations differ are not directly observable. Existing work has begun to investigate the interpretability of embedding dimensions and latent semantic structure \parencite[]{senel_semantic_2018}, offering potential avenues for future methodological extensions.

Second, embedding spaces may not behave uniformly across domains. Embedding spaces are learned from language data and may exhibit non-uniform semantic resolution \parencite{valentini_investigating_2023, wang_unsupervised_2023}, such that domains more extensively represented in training corpora may be encoded at finer granularity than others. This means that equal cosine dissimilarities across concepts may not correspond to equal degrees of representational difference, especially affecting comparisons of $S_b$ across concepts. To mitigate this, researchers may focus on comparisons of concepts from the same semantic domain or include statistical controls, such as general word frequency norms.

Third, sentence encoders are pre-trained on large text corpora that are predominantly English and reflect the cultural contexts of their production. As such, they exhibit documented biases along dimensions such as gender, ethnicity, and sociolect \parencite[e.g.,][]{caliskan_semantics_2017}, and represent meaning within a normatively laden Western reference frame that may be subject to systematic distortions. Particular caution is warranted when investigating normatively or culturally sensitive concepts, where distortions are both more likely and more consequential.

\subsection{Ethical Implications}

Several additional ethical implications warrant consideration and should guide future applications of the framework.

First, it is important to keep in mind that all quantitative claims about concept properties are population-relative. The demonstrations presented in the study draw on SWOW data, in which Western participants are strongly overrepresented. Estimates of sharedness, variance structure, and deviation therefore reflect a skewed population baseline. Treating such estimates as universal properties obscures cultural variation, and representative claims require informed sampling.

Second, when researchers collect custom data, they should account for the heightened sensitivity of open-ended responses relative to closed-ended survey formats. Such responses may contain personal information, reveal identifiable linguistic styles, and, particularly when combined with demographic variables, increase the risk of re-identification \parencite[e.g.,][]{narayanan_robust_2008}. This warrants mitigation strategies such as limiting, aggregating, or removing accompanying personal information to reduce re-identification risk.

Finally, the framework translates individual concept representations into quantitative indices. This is not ethically neutral. Quantification enables comparison, ranking, and sanction, and measures developed in scientific contexts may migrate into downstream decision-making applications such as screening, clinical assessment, or educational tracking, for which they were not designed \parencite[e.g.,][]{kosinski_private_2013, obermeyer_dissecting_2019}. These concerns are particularly important when individuals are the focal unit of analysis. Researchers therefore bear a responsibility that extends beyond scientific utility to the social consequences of measurement itself.

\section{Conclusion}
This paper presented a measurement framework for modeling concept representations and analyzing their interindividual variation. Methodologically, it contributes to a broader trend in computational social science toward modeling standardized textual traces to access mental structures \parencite{aeschbach_mapping_2025,petrenco_centroid_2026}, which may offer a complement to closed-ended psychometric formats such as surveys. 

Substantively, concept representations have explanatory potential across disciplinary boundaries, and the present paper offers only an initial demonstration of how this can be studied empirically. Researchers may apply the framework to directed disciplinary hypotheses, broader and theoretically informed concept sets, and additional statistical controls.

\printbibliography

\end{document}